\documentclass[conference]{IEEEtran}
\IEEEoverridecommandlockouts

\usepackage{url}            

\usepackage{cite}
\usepackage{amsmath,amssymb,amsfonts,bm}
\usepackage{dsfont}
\usepackage{algorithmic}
\usepackage{graphicx}
\usepackage{textcomp}
\usepackage{xcolor}
\usepackage{gensymb}
\usepackage[font=footnotesize]{subcaption}
\usepackage{todonotes}

\usepackage{comment}

\usepackage{etoolbox}

\newtoggle{anonymous}
\settoggle{anonymous}{true} 

\newtoggle{comments}
\settoggle{comments}{true} 
\iftoggle{comments}{
  \newcommand {\alberto}[1]{{\color{orange}{~Alberto: #1}\normalfont}}
  \newcommand {\bjin}[1]{{\color{blue}{~Baihong: #1}\normalfont}}
  \newcommand {\yuxin}[1]{{\color{violet}{~Yuxin: #1}\normalfont}}
  \newcommand {\yingshui}[1]{{\color{green}{~Yingshui: #1}\normalfont}}
  \newcommand {\alex}[1]{{\color{cyan}{~Alex: #1}\normalfont}}
  \newcommand {\red}[1]{{\color{red}{#1}\normalfont}}
}{
  \newcommand {\alberto}[1]{{}}
  \newcommand {\bjin}[1]{{}}
  \newcommand {\yuxin}[1]{{}}
  \newcommand {\alex}[1]{{}}
  \newcommand {\yingshui}[1]{{}}
  \newcommand {\red}[1]{{}}
}

\usepackage{acronym}
\acrodef{SVM}{Support Vector Machine}
\acrodef{AE}{AutoEncoder}
\acrodef{RNN}{Recurrent Neural Network}
\acrodef{CNN}{Convolutional Neural Network}
\acrodef{OC-SVM}{One-class Support Vector Machine}
\acrodef{CPS}{Cyber-Physical System}
\acrodef{FDD}{Fault Detection and Diagnosis}
\acrodef{LDA}{Linear Discriminant Analysis}
\acrodef{PCA}{Principal Component Analysis}
\acrodef{MSE}{Mean Square Error}
\acrodef{LAMPS}{Laser Additive Manufacturing Pilot System}
\acrodef{AM}{Additive Manufacturing}
\acrodef{SLS}{Selective Laser Sintering}
\acrodef{IR}{infrared}
\acrodef{IoT}{Internet of Things}
\acrodef{ROC}{Receiver Operating Characteristic}
\acrodef{RMS}{Root-Mean-Square}
\acrodef{AUC}{Area Under the Curve}

\newcommand{\Score}{f_\text{score}}

\newcommand{\expctover}[2]{\mathbb{E}_{#1}\!\left[#2\right]}
\newcommand{\unit}[1]{\mathds{1}\left\{#1\right\}}

\def\BibTeX{{\rm B\kern-.05em{\sc i\kern-.025em b}\kern-.08em
    T\kern-.1667em\lower.7ex\hbox{E}\kern-.125emX}}
\begin{document}

\title{An Encoder-Decoder Based Approach for\\ Anomaly Detection with Application in Additive Manufacturing
{
\thanks{This work is supported in part by the National Research Foundation of Singapore through a grant to the Berkeley Education Alliance for Research in Singapore (BEARS) for the Singapore-Berkeley Building Efficiency and Sustainability in the Tropics (SinBerBEST) program, and by the National Science Foundation under Grant No.~1645964. BEARS has been established by the University of California, Berkeley as a center for intellectual excellence in research and education in Singapore.}
\thanks{$^\dagger$Baihong~Jin and Yingshui~Tan contributed equally to this paper.}
\thanks{$^*$Alexander~Nettekoven prepared and processed the experimental data. Dr.~Yuxin~Chen contributed to the theoretical aspects of this paper.}
}}

\author{Baihong~Jin$^{\dagger1}$~~Yingshui~Tan$^{\dagger1,2}$~~Alexander~Nettekoven$^{*3}$~~{Yuxin Chen}$^{*4}$~~{Ufuk Topcu}$^{3}$\\~~{Yisong Yue}$^{4}$~~{Alberto~Sangiovanni-Vincentelli}$^{1}$\\
$^{1}$Department of EECS, University of California, Berkeley, USA\\
$^{2}$RWTH Aachen University, Germany\\
$^{3}$University of Texas at Austin, USA\\
$^{4}$California Institute of Technology, USA
}

\maketitle

\begin{abstract}
We present a novel unsupervised deep learning approach that utilizes the encoder-decoder architecture for detecting anomalies in sequential sensor data collected during industrial manufacturing. Our approach is designed not only to detect whether there exists an anomaly at a given time step, but also to predict what will happen next in the (sequential)  process. We demonstrate our approach on a dataset collected from a real-world \ac{AM} testbed. The dataset contains \ac{IR} images collected under both normal conditions and synthetic anomalies. We show that the encoder-decoder model is able to identify the injected anomalies in a modern \ac{AM} manufacturing process in an unsupervised fashion. In addition, it also gives hints about the temperature non-uniformity of the testbed during manufacturing, which is what we are not aware of before doing the experiment.  
\end{abstract}

\begin{IEEEkeywords}
additive manufacturing, anomaly detection, fault detection
\end{IEEEkeywords}

\section{Introduction}

Anomaly detection is an important technique that serves as the basis of applications across a diverse variety of domains, such as fault detection, intrusion and fraud detection~\cite{phua2010comprehensive}, and process control. The goal of anomaly detection is to identify patterns in data that do not conform to a well-defined notion of normal behavior~\cite{chandola2009anomaly}. Early detection of anomalies and faults allows us planning preventive maintenance for model manufacturing, and thus it is crucial for process control.
 
The availability of massive amount of data due to the introduction of pervasive sensing techniques has brought plenty of opportunities for data-driven anomaly detection applications; however, an unresolved challenge is how to make use of these data for anomaly detection, especially when there is no label information that can be used to differentiate between normal and anomalous working conditions.

\subsection{Learning-Based Anomaly Detection}
Depending on the availability of labeled anomalous data, learning-based anomaly detection approaches can generally be categorized into \textit{supervised} and \textit{unsupervised} methods. Supervised methods utilize label information for both normal and anomalous data to train classification models. The trained classification models from supervised learning can not only tell the existence of faults but also indicate the likelihood of an input belonging to a particular type of fault.

A review of the literature reveals that data-driven approaches relying on supervised learning have demonstrated promising results in various applications, e.g.~\ac{FDD} in air conditioning systems~\cite{li2016data,li2016fault,jin2019detecting}. 

To train a well-performing model using supervised learning, a good amount of labeled data from both normal and anomalous conditions are needed, which is not always easy to obtain in practice. 

In addition, supervised models typically lack the ability to identify an unseen example that does not belong to any of the classes that appear in the training set. In the context of anomaly detection, models trained with supervised learning are likely to give incorrect predictions on out-of-distribution data instances. This is a limitation of supervised methods because it is almost impossible to obtain every possible type of anomaly that could happen on a system. To address this problem, Jin~et~al. recently proposed a \ac{FDD} method that uses Monte-Carlo dropout~\cite{jin2019detecting} to estimate the prediction uncertainty of deep neural networks. The method was applied to the identification of incipient faults that are not represented in the training data that only consists of labeled data of normal and severe faults.

In scenarios where labeled \emph{anomalous} data are scarce or unavailable, unsupervised and semi-supervised anomaly detection approaches are usually applied, because only normal data are required to train a detection model. The two approaches differ in their assumptions about the labels of training data. In semi-supervised learning, it is assumed that the training set is comprised of only data instances from the normal class\footnote{Note that semi-supervised anomaly detection differs from the traditional notion of ``semi-supervised learning'' in machine learning, where both label and unlabeled data are used simultaneously for training.}, while in an unsupervised setting, it is often implicitly assumed that few anomalous instances can exist in the training data~\cite{chandola2009anomaly}. We note that the approach we introduce in this paper can apply to both settings. We choose to use the term ``unsupervised learning'' throughout this paper to refer to both situations where normal data account for the majority or the entirety of the training data.  Although unsupervised approaches usually lack the discriminative ability to assign labels to anomalous data, it is still considered an appealing complement to supervised approaches in many real-world applications.

Recently, neural network approaches, especially deep neural networks, have attracted much attention from the machine learning community, 
due to their ability to process natural data in their raw form and learn internal representations that can be used for detecting or classifying patterns~\cite{lecun2015deep}. Yet, as the authors of the recent review paper~\cite{lecun2015deep} also pointed out, supervised learning accounted for the majority of the recent success of deep learning, while unsupervised learning is expected to be far more important in the longer term. This paper aims at taking advantage of the recent development of deep learning and provide a methodology for developing unsupervised anomaly detection algorithms for handling sequential sensing data in industrial applications.

\subsection{Our Contributions}

In particular, we investigate the applicability of an encoder-decoder approach on sequential image sensing data collected in a real industrial setting. The contributions of this paper are two-fold:
\begin{itemize}
    \item We propose using an encoder-decoder architecture for detecting anomalies in sequential image sensing data collected from \ac{AM} process. The learning process is unsupervised, meaning that no anomalous data are needed \textit{a priori} to train the detection model. 
    \item We design a \ac{CNN}-based encoder-decoder network to monitor the manufacturing process of the \ac{LAMPS} testbed, a platform that uses \ac{SLS} technology for \ac{AM}. In our experiment, the network can not only detect the artificially injected laser anomalies with high accuracy, but also can indicate regions of the manufacturing testbed where the temperature is higher than usual. Our results demonstrate the effectiveness of the proposed algorithm in detecting anomalous phenomena.
\end{itemize}

\subsection{Paper Organization}

The remainder of this paper is organized as follows. In Sec.~\ref{sec:related}, we will give the background about \ac{LAMPS}, the encoder-decoder architecture and deep-learning-based anomaly detection approaches. We will define the anomaly detection problem for sequential data in Sec.~\ref{sec:problem-definition}. We will describe our anomaly detection methodology for sequential image data in Sec~\ref{sec:methodology}. In Sec.~\ref{sec:algorithmic-details}, we will describe in details our anomaly detection algorithm when applied to a real-world \ac{AM} dataset with injected faults. Experimental results will be demonstrated and evaluated in Sec.~\ref{sec:experiment}. We will discuss future work and conclude the paper in Sec.~\ref{sec:conclusion}. 
\section{Background}\label{sec:related}

\subsection{\acf{LAMPS}}

\ac{AM} technologies have transformed the manufacturing landscape.~\cite{huang_review} In contrast to traditional manufacturing technologies, 
\ac{AM} technology is capable of printing 3D parts with highly complex geometries in a single process step. 
Due to its versatility, \ac{AM} technology are used in a wide variety of applications such as medical devices and aircraft manufacturing~\cite{tofail_review}. One of the prominent \ac{AM} technologies is Selective Laser Sintering (\ac{SLS}) that uses a laser to form solid parts out of powdered material. Building parts with consistent high-quality is a key challenge for the \ac{SLS} process today~\cite{huang_review}.
Therefore, having an algorithm that can monitor the \ac{SLS} printing process and can indicate potential anomalies will significantly improve \ac{SLS} process control. This, in turn, will lead to improved part quality and ensure repeatability.

We now briefly introduce the \ac{SLS} printing process and the testbed we used for data collection and testing purposes. \ac{SLS} utilizes a laser to fuse powder geometries layer-by-layer and hereby generates a solid 3D structure. At the beginning of each layer, a roller spreads a new powder layer across the powder bed. Once the powder has been spread, the laser melts the cross-section of the desired part according to the digital 3D model. After the laser has finished scanning for the current layer, a new powder layer is spread and the scanning process is repeated. Over time the melted powder locations on each layer will cool down and will solidify to one.

\begin{figure}[tb]
    \begin{subfigure}[t]{0.44\linewidth}
    \centering
    \includegraphics[height=3cm]{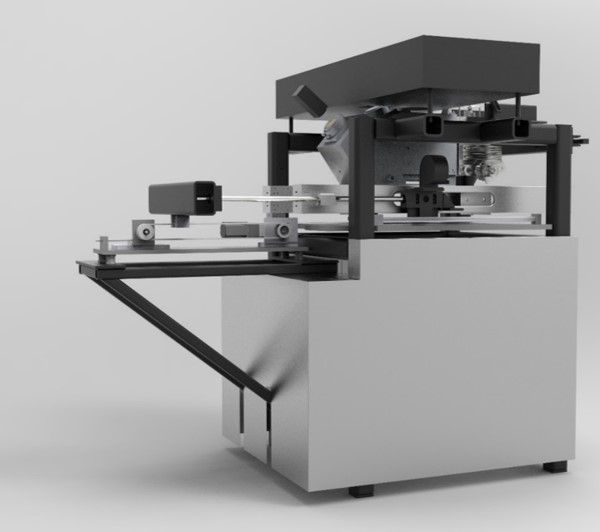}
    \caption{LAMPS architecture}
    \label{fig:LAMPS_architecture}
    \end{subfigure}
    \begin{subfigure}[t]{0.55\linewidth}
    \centering
    \includegraphics[height=3cm]{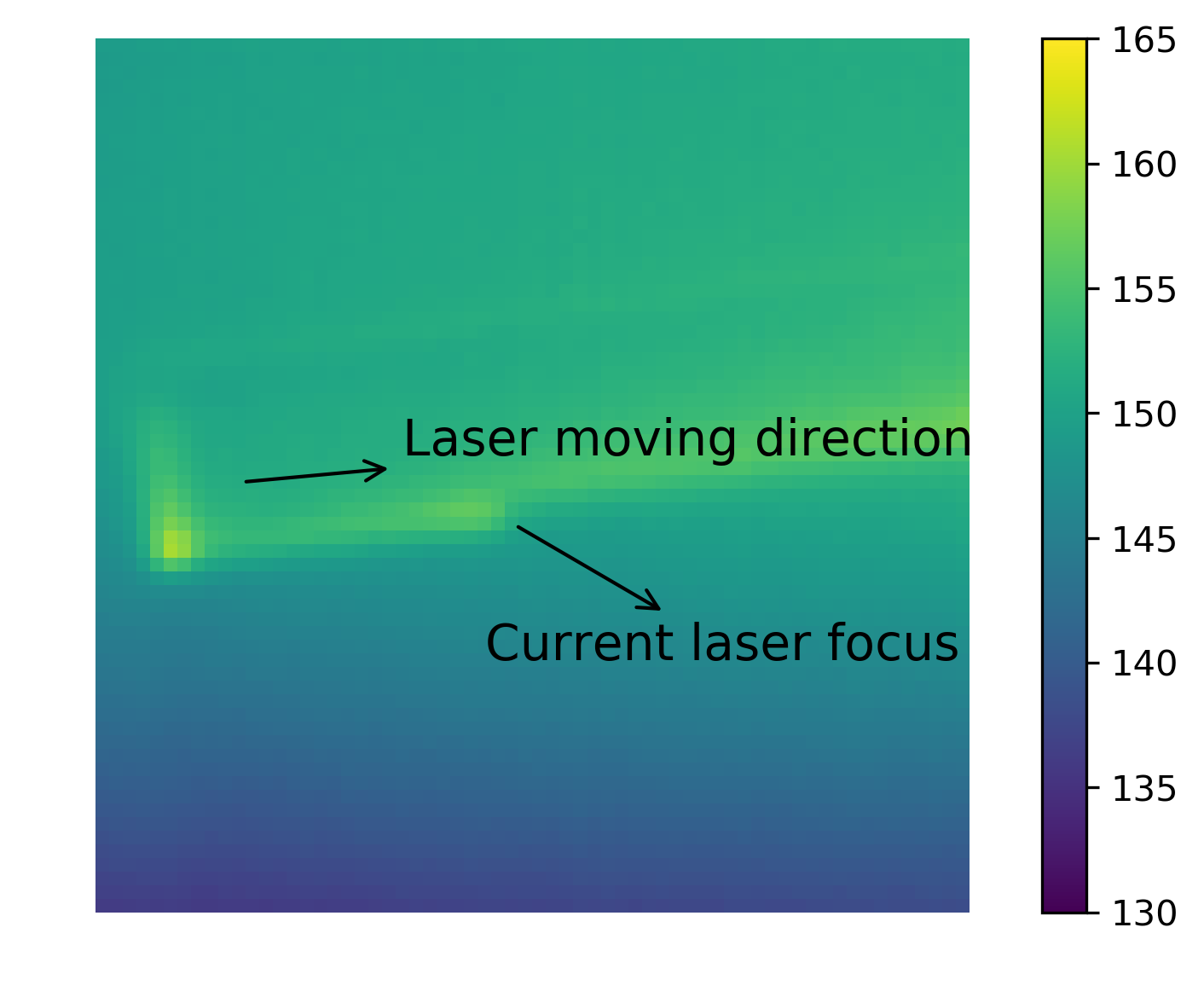}
    \caption{Boresight image example}
    \label{fig:boresight_example}
    \end{subfigure}
    \caption{(a) \ac{LAMPS} testbed~\cite{phillips_thesis}. (b) An example image captured by the boresighted temperature sensor.}
\end{figure}

\ac{LAMPS} is a \ac{SLS} testbed that was designed and built for process control research. LAMPS is capable of building 3D parts out of high-performance plastics (melting temperatures as high as $350$~\textdegree{}C) and is equipped with a variety of sensors, such as \ac{IR} and visual cameras, that provide \textit{in-situ} measurement access. Fig.~\ref{fig:LAMPS_architecture} shows the general architecture of the \ac{LAMPS} testbed.

In the context of this paper, we focus on the high-speed mid-wave infrared (\ac{IR}) camera which is bore-sighted with the laser optics to record the laser focus and its immediate surrounding. The camera has a resolution of $64\times64$ pixels and has a maximum recording frame rate of $2.24$~kHz. Fig.~\ref{fig:boresight_example} shows an exemplary \ac{IR} image of the bore-sighted camera. The recorded IR information is translated to gray-scale (single-channel) temperature images, where the intensity value of each pixel represents the measured temperature value at that pixel.

\subsection{Encoder-decoder architecture}\label{sec:enc-dec-arch}

The encoder-decoder architecture has proven to be a useful approach for learning (deep) representations, 
and is widely used in various application domains of deep learning, including machine translation~\cite{cho2014learning}, and image denoising~\cite{vincent2008extracting}. An encoder-decoder model generally consists of three parts: the encoder, the latent space representation, and the decoder. The purpose of the encoder network $Enc$ is to transform the input data into a latent space representation $\bm{z}$ that is often a vector; the decoder network $Dec$ then produces the output by decoding $\bm{z}$. During training, the encoder and the decoder are trained together to minimize the empirical risk. 

Proper design of the latent space representation $\bm{z}$ is crucial to the successful application of the encoder-decoder approach. Let us take the basic autoencoder model as an example. An autoencoder is a neural network model that is trained to reconstruct its input. In other words, an autoencoder is trained to learn an identity function for the data distribution. By constraining $\bm{z}$ to be a low dimensional vector, the training process encourages the model to learn the most useful information for reconstructing the input.

\subsection{Unsupervised anomaly detection with deep learning}\label{sec:enc-dec-arch}

Supervised deep learning has been extensively studied in various applications domains. 

In fault/anomaly detection tasks, we often do not have access to the entire spectrum of off-nominal data, as well as the labels that come along with it. As a result, unsupervised approaches that do not require labeled anomaly data are more suitable in such scenarios.

In this paper, we aim to explore unsupervised anomaly detection using a deep learning approach. Specifically, we will adopt the encoder-decoder scheme described earlier in Sec.~\ref{sec:enc-dec-arch}.

The general idea behind unsupervised anomaly detection approaches is to find an approximate model that can capture the normal behavior of complex systems. The approximate model can then be used to flag anomalies if the deviation of the predicted behaviors of the trained model from the actual observation exceeds some certain threshold. Examples that share this general idea include  \ac{OC-SVM}~\cite{scholkopf2001estimating,heller2003one,jin2019one}, \ac{PCA}~\cite{lee2012anomaly} and autoencoders~\cite{sakurada2014anomaly}.

The encoder-decoder schemes for anomaly detection that appeared in literature in general fall into three categories, which differ in their prediction outputs: 1) autoencoder models~\cite{malhotra2016lstm}, 2) prediction models, and 3) composite models~\cite{srivastava2015unsupervised} that performs both reconstruction and regression. We denote our observation at time instant $\tau$ by $S_\tau$. The observations we observe in time then forms as a sequence $\{S_\tau\}$. Let $g$ be a function that maps an input sequence of length $p$ to an output sequence of length $q$. These encoder-decoder schemes are therefore summarized below:
\begin{align}\label{eqn:enc-dec-formulation}
    &(S_{\tau_{-p+1}},\ldots,S_{\tau_0})\notag\\
    &\quad\xrightarrow{g}\begin{cases}
        (S_{\tau_{-p+1}},\ldots,S_{\tau_0}),~\text{reconstruction model},\\
        (S_{\tau_{1}},\ldots,S_{\tau_q}),~\text{regression model},\\
        (S_{\tau_{-p+1}},\ldots,S_{\tau_q}),~\text{composite model}.
    \end{cases}
\end{align}

As previously described, reconstruction models (a.k.a.~autoencoders) aim to find a compact representation for input data distribution. Depending on the format of the input data, different neural network architectures or their combinations are used to design encoders and decoders. Autoencoders are first trained on data that are normal or almost fault-free. The reconstruction errors given by autoencoder models are often used as anomaly scores to indicate potential anomalies. This approach is seen in previous literature for anomaly detection in multivariate timeseries~\cite{malhotra2016lstm}.  

Similarly, we can also use the encoder-decoder architecture for prediction tasks. In the case of time series data, a neural network prediction model can be trained to predict the future from past observations. Taking the past $p$ observations as input $(S_{\tau_{-p+1}},\ldots,S_{\tau_0})$, the model is trained to predict the next $q$ observations $(S_{\tau_{1}},\ldots,S_{\tau_q})$. During training, the encoder will look for information needed for the decoder to predict the future, and encode the information as latent space representations. In this case, the prediction errors are used to indicate potential anomalies. 

The authors of~\cite{srivastava2015unsupervised} argue that a composite model, by performing the reconstruction and the regression tasks simultaneously, can overcome the drawbacks of each one when performed alone, and thus achieving better performance at learning useful representations in the data. Previous literature reports on schemes for detecting anomalies in videos~\cite{medel2016anomaly} and multivariate time series. In our case study to be later discussed, we designed our encoder-decoder model as a composite model to leverage the advantages of both reconstruction and regression models.

\section{The Anomaly Detection Problem}\label{sec:problem-definition}
Assume that we are given a series of observation data, 
$S_{\tau_0},S_{\tau_1},\ldots,S_{\tau_i},\ldots$, where each $S_{\tau_i} \in \mathcal{S}$ ($\mathcal{S}$ being the input domain) 
denotes the representation of the $i$th data point in the sequential data. In the anomaly detection setting, we assume that all data points from the training set are in the normal state. 

Let $\mathcal{F}$ be a model class, where each $f \in \mathcal{F}: \mathcal{S} \rightarrow \mathbb{R}_{\geq 0}$ denotes a score/fitness function that characterizes how close a data point is to an abnormal state, i.e., larger $f$ implies higher chance of a data point being abnormal.

For a given threshold value $\epsilon > 0$, we define the detection precision of $f$ as
\[
\text{prec}(f,\epsilon) = \expctover{S}{\unit{S\text{~is abnormal}} \mid {f(S)>\epsilon}}
\]
where the expectation is taken over the distribution of the test data, and the recall of $f$ as
\[
\text{recall}(f,\epsilon) = \expctover{S}{\unit{f(S)>\epsilon} \mid S \text{~is abnormal} }
\]
Our goal is to learn a score function $\Score \in \mathcal{F}$ and a corresponding threshold $\epsilon$, such that $(\Score,\epsilon)$ achieves the best detection accuracy and recall of anomalies on the (unseen) test data.

\section{Methodology}\label{sec:methodology}
 
We utilize the encoder-decoder architecture described in Sec.~\ref{sec:enc-dec-arch} to design a neural network that can be used to detect possible anomalies in \ac{AM} process. Since we are dealing with image data in \ac{LAMPS} application, we choose to use \acp{CNN}~\cite{krizhevsky2012imagenet} as the main building blocks for our encoder-decoder model. Our approach uses the composite prediction model described earlier in Sec.~\ref{sec:enc-dec-arch} -- the designed model will not only attempt to reproduce the input but also predict what will happen next.

\subsection{CNN-based encoder-decoder model}

In our unsupervised learning setting, we only have access to data points collected under normal condition. The learning goal is to use a neural network to model the normal behaviors of the system under study. Outliers to the learned distribution will be identified as potential anomalies.

Let us suppose that each observation $S_\tau$ in the sequential data is a single-channel 2D image of dimension $m\times n$, i.e.~$S_\tau\in\mathbb{R}^{m\times n}$. To capture the temporal correlations among the observations, a sliding window approach can be used to divide the original image sequence into \textit{snippets}, where each snippet $\mathbf{Z}_k\in\mathbb{R}^{m\times n\times (p+q)}$ comprise of $p+q$ consecutive frames, and $k$ is the index of the snippet.

When a regression or composite prediction scheme is used to train an encoder-decoder model, the frames in a snippet constitute the input and the output. For training a regression model, the first $p$ frames in a snippet $\mathbf{X}_k\in\mathbb{R}^{m\times n\times p}$ constitute the model input, and the rest $q$ frames are the output to be predicted. In the case of a composite model, the input is still the $\mathbf{X}_k$, and the output is the entire $p+q$ frames. If we view the frames in a snippet as channels in an image, the learning problem can be cast as an \textit{image-to-image} translation task. To be more specific, we will train the encoder-decoder network $\mathcal{M}$ to learn a mapping $g: \mathbb{R}^{m\times n\times p} \rightarrow \mathbb{R}^{m\times n\times (p+q)}$ that transforms a $p$-channel image input $\mathbf{X}_k$ to an output $\hat{\mathbf{Z}}_k$ with $(p+q)$ channels. 
The prediction output $\hat{\mathbf{Z}}_k$ can be seen as the combination two parts, $\hat{\mathbf{X}}_k$ and $\hat{\mathbf{Y}}_k$. $\hat{\mathbf{X}}_k$ is the reconstruction of the $p$ input frames, and $\hat{\mathbf{Y}}_k$ is a prediction of the $q$ frames following the input frames. 

When training the encoder-decoder model, we aim to minimize the errors on both the reconstruction part and the regression part. Since the model input and output are both images, the following pixel-wise \ac{MSE} can be used as the error metric on frame $S_\tau$.
\begin{align}\label{eqn:raw-error}
    \ell_{mse}(S_\tau, \hat{S}_\tau) = \Vert{S_\tau-\hat{S}_\tau}\Vert_F,
\end{align}
where $\Vert{\cdot}\Vert_F$ is the Frobenius norm of a matrix.

Let us suppose the frames in snippet $k$ are taken at time instants $\tau_k^0, \tau_k^1, \ldots, \tau_k^{p+q-1}$. By choosing~\eqref{eqn:raw-error} as the error metric, we can define the reconstruction error $e_k^\text{rec}$ and regression error $e_k^\text{reg}$ on snippet $k$ as follows 
\begin{align}\label{eqn:error-def}
    e_k^\text{rec}&\doteq\sum_{0 \leq i < p}\ell_{mse}(S_{\tau_k^i}, \hat{S}_{\tau_k^i}),\\
    e_k^\text{reg}&\doteq\sum_{0 \leq i < p+q}\ell_{mse}(S_{\tau_k^i}, \hat{S}_{\tau_k^i}).
\end{align}
The loss function $L$ to minimize during model training can then be defined as as the weighted sum of reconstruction error $e_k^\text{rec}$ and regression error $e_k^\text{reg}$ on all training samples $k\in\mathcal{K}$.
\begin{align}\label{eqn:loss-function}
    &L = \sum_{k} e_k^\text{rec} + \lambda e_k^\text{reg}
\end{align}
where $\lambda$ is a weighting factor that adjusts the relative importance between the reconstruction error and the regression error.

\subsection{Using the trained model for anomaly detection}

Assuming the trained encoder-decoder model has learned a good representation of the normal behavior of the system, the differences between the predicted images and their corresponding ground truth can be used to indicate possible anomalies. By comparing the images, we are essentially getting a large number of pixel-wise errors, and thus a method is needed to process this information in order to detect and locate the anomalies.

One simple idea is to use the original loss function~\eqref{eqn:loss-function} that we used for training the network. These loss values can be derived directly from the prediction results, and can be used as good indicators for evaluating the network's prediction quality; however, this approach also suffers from two drawbacks. First, if the anomaly is only localized to a small area, it is likely that the prediction errors are only significant in a small part of the image. When we calculate the pixel-wise \ac{MSE} over the entire image, useful indications of anomalies may be buried in noise and averaged out. In addition, even if a significant loss is observed on an image, this approach only indicates a potential anomaly on the image level, but it does not give further hint about the occurrence of this anomaly. It is unknown whether the anomaly is local to only a small area or affects the entire image.

\begin{figure*}[tb]
    \centering
    \includegraphics[width=0.8\linewidth]{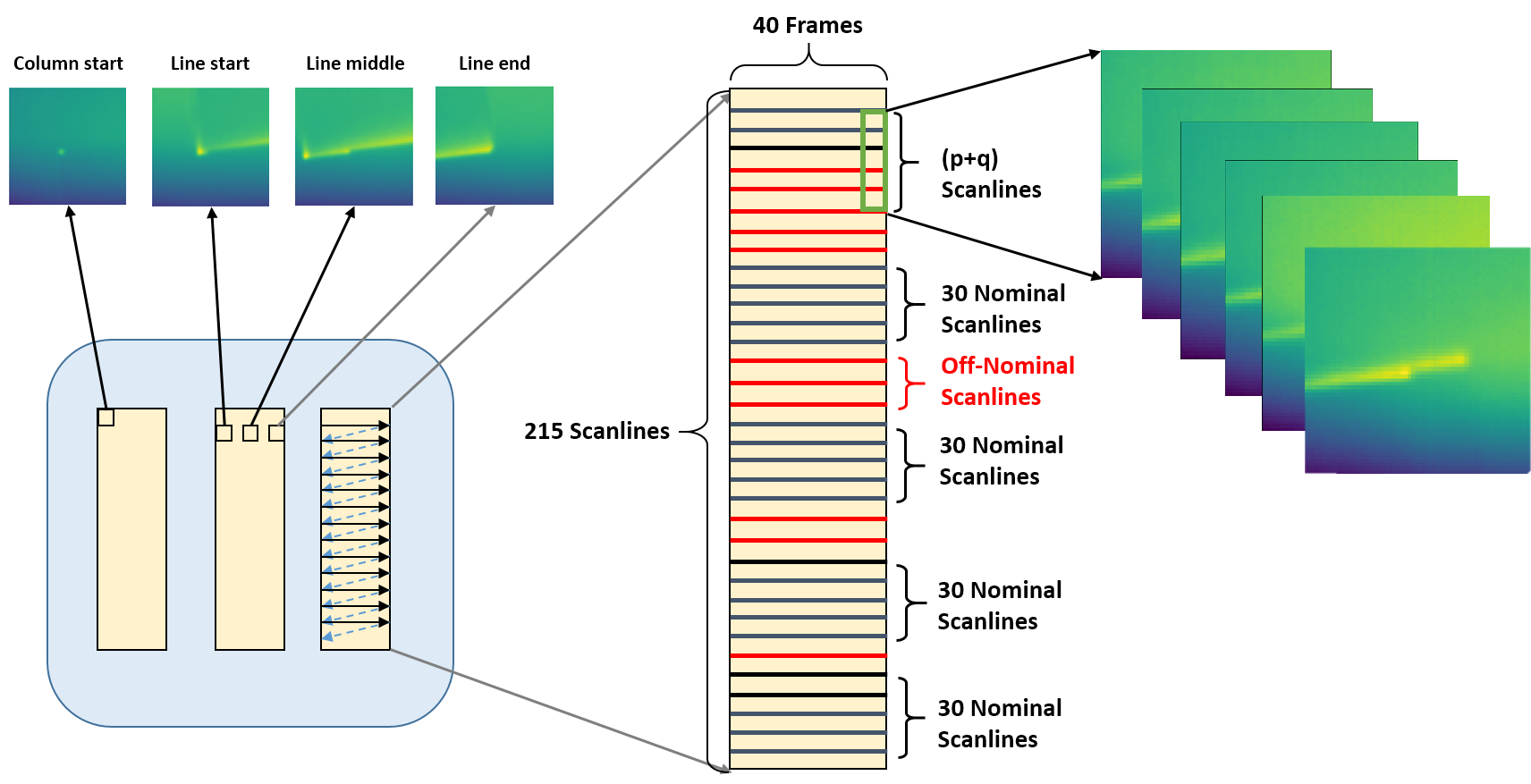}
    \caption{An top-down illustration of the benchmark dataset. (Left) On the top we show examples of boresight images when the laser head it at column start, line start, line middle and line end, respectively; on the bottom we illustrate the trace of scanning laser; (Middle) the nominal and off-nominal scan lines pattern; (Rigth) Examples of preprocessed data (boresight images)}
    \label{fig:rectangles}
\end{figure*}

To address the above mentioned challenge when using for evaluating image-to-image differences, we propose a ``spatial scoping'' approach: we aim to find a $a\times b$ window from the $m\times n$ error matrix $E_\tau={S}_\tau-\hat{S}_\tau$ that has the maximum Frobenius norm. The new error metric $\ell_{ss}$ is thus defined as:
\begin{align}\label{eqn:spatial-scoping-loss}
    \ell_{ss}(S_\tau, \hat{S}_\tau) = \max_{\substack{0\leq i \leq m-a \\ 0\leq j \leq m-b}} \Vert{E_\tau^{i:i+a,j:j+b}}\Vert_F,
\end{align}
where the superscript in $E_\tau^{i:i+a,j:j+b}$ indicates the position of the window in the original error matrix $E_\tau$.

Next we define two anomaly scores, the \textit{reconstruction anomaly score} $\Score^\text{rec}(\tau)$ and the \textit{regression anomaly score} $\Score^\text{reg}(\tau)$, as the metrics for evaluating the ``degree of anomaly'' of an observation $S_\tau$. Note that there is more than one snippet that encompasses $S_\tau$ because we used a sliding window approach to generate the snippets. To get a single anomaly score taking into account the prediction errors from all relevant snippets, we define the anomaly score as the average prediction errors from all these snippets. 

Suppose $K_\tau^\text{rec}$ is the index set of snippets whose reconstruction window covers $S_\tau$, and $K_\tau^\text{rec}$ is the index set of snippets whose reconstruction or regression window covers $S_\tau$. The reconstruction and regression anomaly scores on $S_\tau$ can be defined by as follows
\begin{subequations}
\begin{align}
    \Score^\text{rec}(\tau) &= \frac{1}{\vert{K_\tau^\text{rec}}\vert}\sum_{k_\in K_\tau^\text{rec}}e_k^\text{rec},~\text{reconstruction}\label{eqn:rec-anomaly-score-def}\\
    \Score^\text{reg}(\tau) &= \frac{1}{\vert{K_\tau^\text{reg}}\vert}\sum_{k\in K_\tau^\text{reg}}e_k^\text{reg},~\text{regression}\label{eqn:reg-anomaly-score-def}
\end{align}
\end{subequations}
where $\vert{K_\tau^\text{rec}}\vert$ and $\vert{K_\tau^\text{reg}}\vert$ are the cardinalities of sets $K_\tau^\text{rec}$ and $K_\tau^\text{reg}$ respectively. A notable difference between $K_\tau^\text{rec}$ and $K_\tau^\text{reg}$ is their sizes. In the reconstruction case, all snippets whose regression windows (of length $p$) cover $S_\tau$ are included in $K_\tau^\text{rec}$. As a result, $\vert{K_\tau^\text{rec}}\vert = p$, except at the start or end of sequence $\{S_\tau\}$ because at the boundaries there are be fewer snippets covering an observation. In the regression case, as long as an anomaly is seen in either the reconstruction window (of length $p$) or the regression window (of length $q$), the anomaly would (probably) be caught in the regression error. Therefore, $\vert{K_\tau^\text{reg}}\vert = p+q$ except at the start or end of sequence $\{S_\tau\}$.  

The anomaly scores introduced above can be used to evaluate how likely an observation $S_\tau$ will correspond to an anomalous state of the system under study. Later in Sec.~\ref{sec:algorithmic-details} and Sec.~\ref{sec:experiment}, we will present a case study on \ac{LAMPS} to illustrate our proposed approach.

\section{Algorithmic Details for LAMPS}\label{sec:algorithmic-details}

\subsection{Benchmark dataset with synthetic faults}\label{sec:benchmark}

Fig.~\ref{fig:rectangles} shows the laser trajectory in LAMPS machine. It is clear that the laser follows a periodical motion pattern in our experiment. The laser firstly moves rightward till the right boundary of the column and then move leftward to the left boundary. During each period of motion, the laser power will move forward 1 unit in the line axis. There is no laser power in the leftward process which is depicted with dashed lines in Fig.~\ref{fig:rectangles}. We therefore only took the rightward process into consideration in this experiment. 

For testing our anomaly detection algorithm, we created an ``off-nominal'' build with the \ac{LAMPS} machine. During this build, the laser power was altered at specific time instances from its nominal power. We will detail the layout of the build below.

The off-nominal build consisted of three columns being built over the course of 250 layers. Each column had the same off-nominal pattern applied in order to create a large dataset. For each layer, the laser scanned the rectangles (the horizontal section of the columns) with straight scan lines that were horizontally aligned. Off-nominal conditions were applied to every fourth layer by scanning specific scan lines with off-nominal laser power instead of nominal laser power. The off-nominal conditions were only applied every fourth layer to ensure that there would be no temperature influences between off-nominal layers.

The horizontal cross-section of each column is of a rectangle shape. 
Fig.~\ref{fig:rectangles} illustrates the off-nominal scan line pattern for one of the three rectangles. Each rectangle consists of 215 horizontal scan lines and the bore-sight camera is able to approximately take 40 frames for each scan line. The off-nominal laser power magnitude stayed the same within every off-nominal layer, but was continuously changed throughout the build. For more comprehensive testing, the anomalies injected have different lasting areas, from 1 line to 4 lines.

\subsection{CNN-based encoder-decoder network design}\label{sec:network-design}

\begin{figure}[tb]
    \centering
    \includegraphics[width=0.8\linewidth]{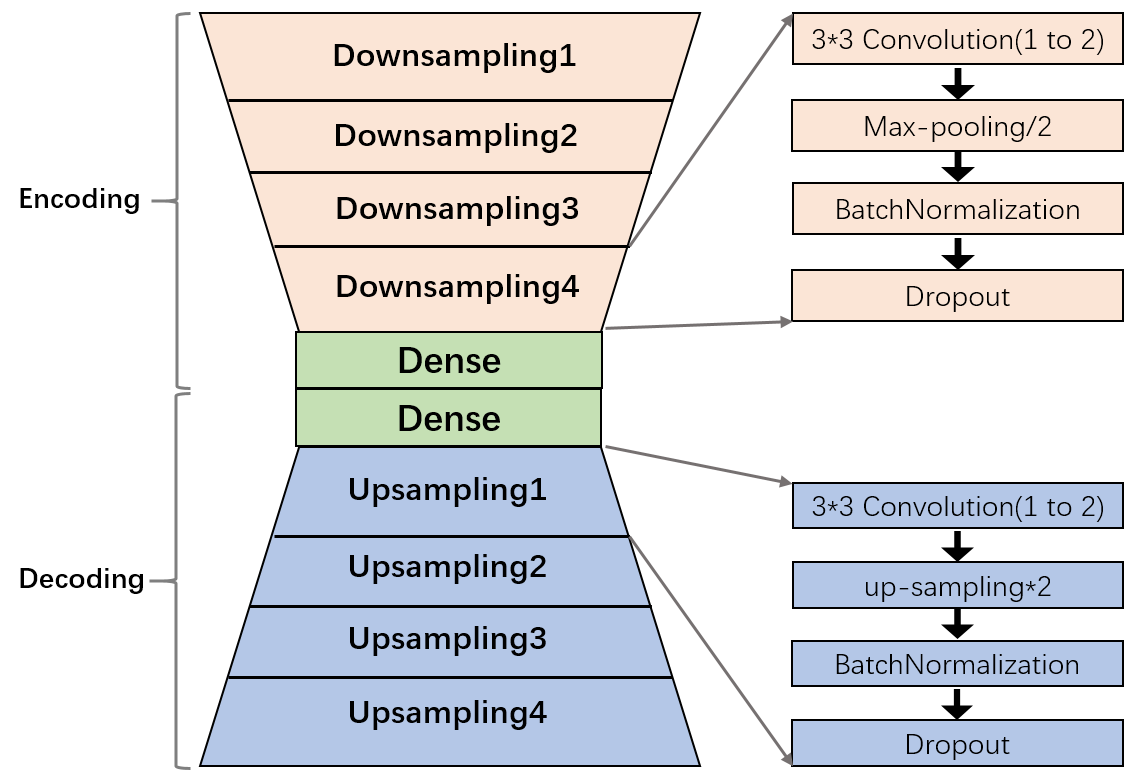}
    \caption{Network Structure}\label{fig:CNN}
\end{figure}

To create training data for the encoder-decoder model, we divide the dataset into snippets with each consisting $p+q$ frames. As illustrated in Fig.~\ref{fig:rectangles}, each snippet is oriented in the vertical direction and spans $p+q$ scan lines. The frames within a scan line are ordered by the time sequence the images were taken. Let us denote by $S_{i,j}$ the $j$th frame taken in the $i$th scan line; here the indices for column number and layer number are omitted for brevity. The encoder-decoder model is trained to transform each input data point $\mathbf{X}_k = (S_{i,j}, S_{i-1,j}, \ldots, S_{i-p+1,j})\in\mathbb{R}^{m\times n\times p}$ into a predicted output $\hat{\mathbf{Z}}_k = (\hat{S}_{i,j}, \hat{S}_{i-1,j}, \ldots, \hat{S}_{i-p+1,j})\in\mathbb{R}^{m\times n\times(p+q)}$.

We choose a VGG-based~\cite{Simonyan2015VeryDC} structure. The convolution kernel size is chosen to be $3\times3$ and the pooling kernel size is chosen to be $2\times2$ to build a deeper network instead of using a large kernel size.
As shown in Fig.~\ref{fig:CNN}, in our network there are four stacked down-sampling layer groups and four stacked up-sampling layer groups to sample the data and reconstruct the data respectively. Each down-sampling group has one or two Convolution layers (depending on the network depth) and a ``Maxpooling'' layer and correspondingly each up-sampling layer has the same number of ``Convolution'' layers and a ``Up-sampling'' layer.
Functionally, when the data is input into the neural network, each down-sampling layer group will down-sample the spatial dimensions (width, height) and double the depth of the data while each up-sampling layer will up-sample the spatial dimensions and halve the depth.
Between the up-sampling groups and down-sampling groups, we set two fully connected layers, from which the latent space representations can be extracted. In addition, in order to prevent the network from over-fitting, we add a ``Dropout'' layer to each group. To make training more efficient, we add a ``BatchNormalization''~\cite{Ioffe2015BatchNA} layer to each group between the ``Maxpooling'' layer and the ``Dropout'' layer.

\section{Experimental Evaluation} \label{sec:experiment}

\begin{figure}[tb]
\centering
    \begin{subfigure}[t]{0.32\linewidth}
    \centering
    \includegraphics[width=0.95\linewidth]{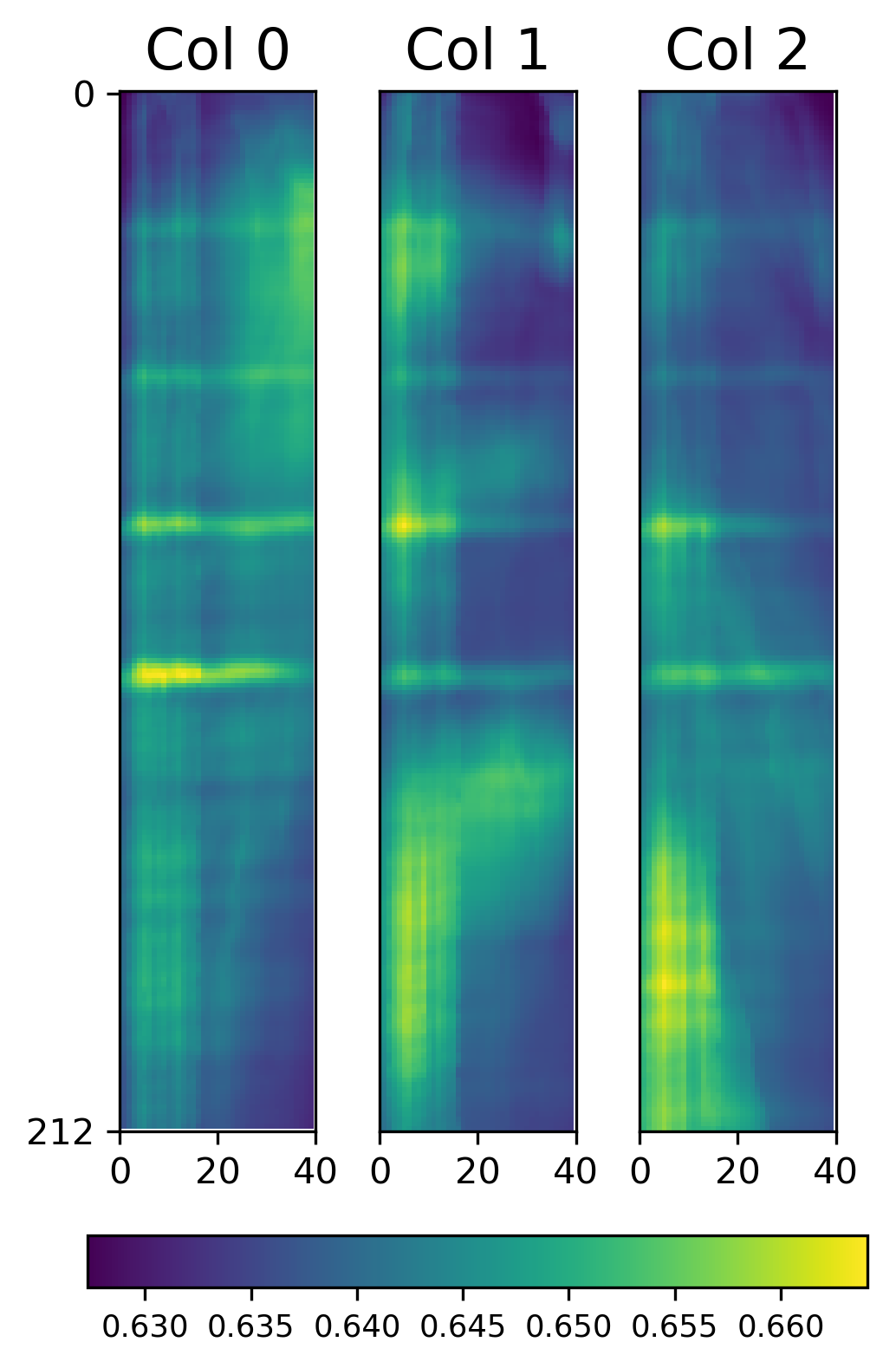}
    \caption{Reconstruction}
    \label{fig:top-down-reconstruction-error}
    \end{subfigure}
    \begin{subfigure}[t]{0.32\linewidth}
    \centering
    \includegraphics[width=0.95\linewidth]{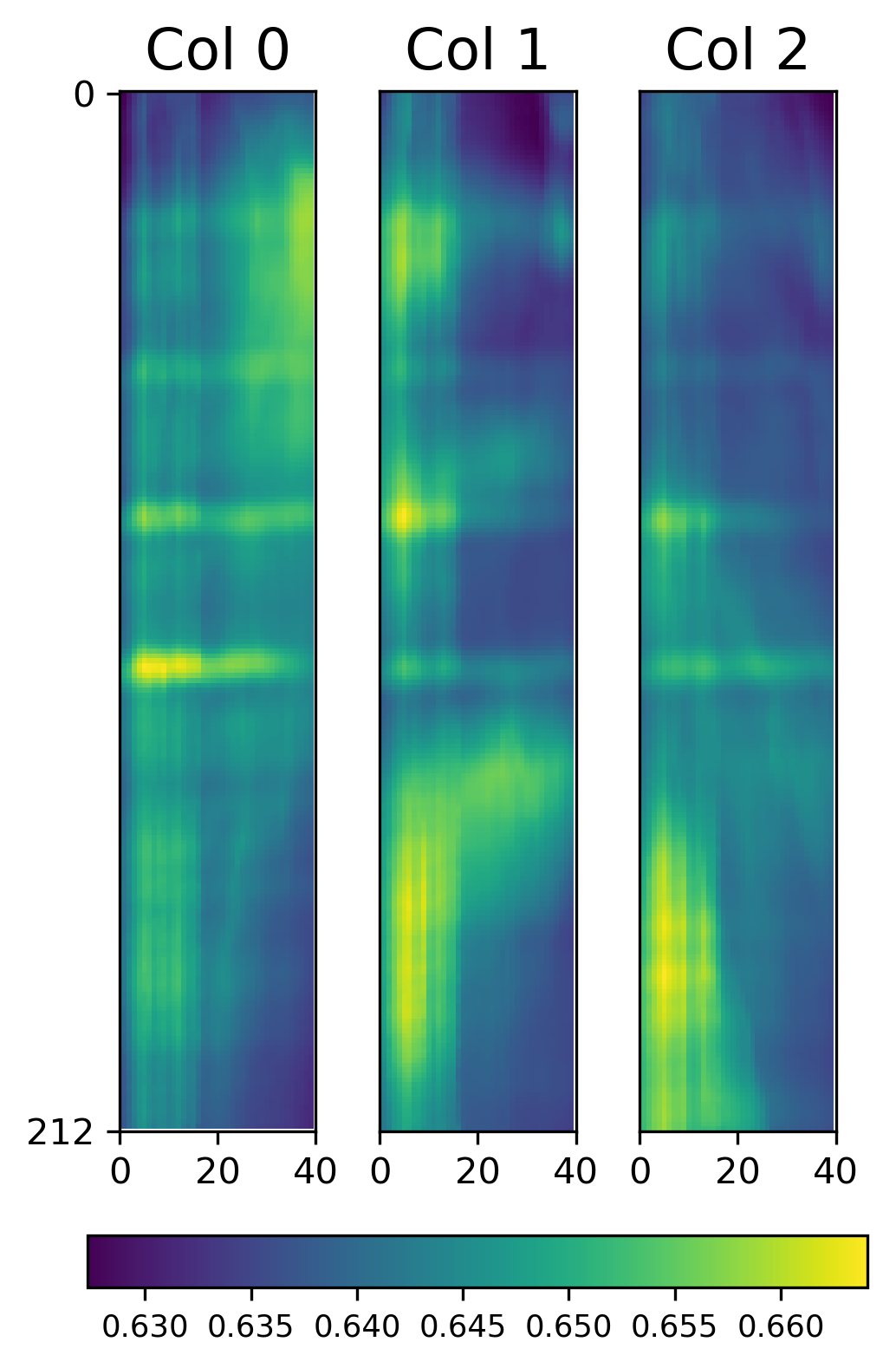}
    \caption{Regression}
    \label{fig:top-down-regression-error}
    \end{subfigure}
    \begin{subfigure}[t]{0.325\linewidth}
    \centering
    \includegraphics[width=0.95\linewidth]{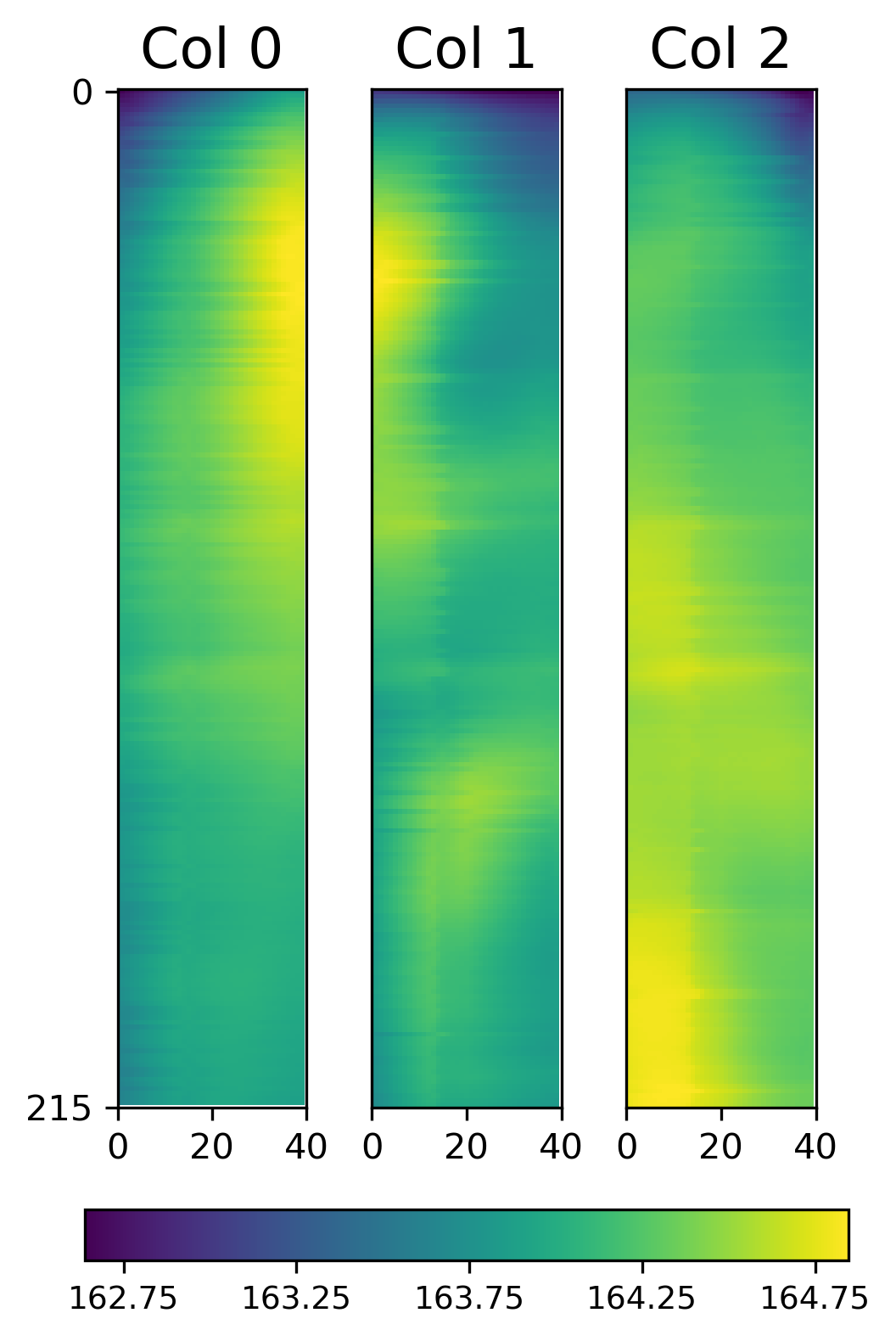}
    \caption{Temperature}
    \label{fig:top-down-temperature}
    \end{subfigure}
\caption{Top-down views}
\label{fig:top-down}
\end{figure}

\begin{figure*}[t]
\centering
    \begin{subfigure}[t]{0.49\textwidth}
    \centering
    \includegraphics[width=8cm]{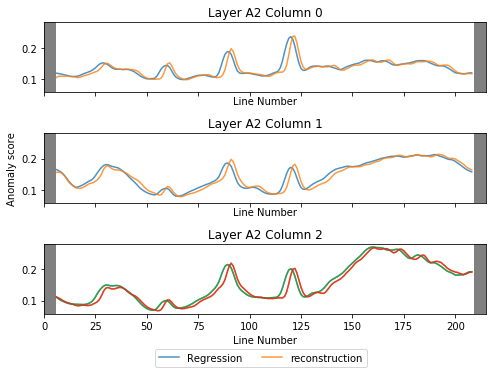}
    \caption{}
    \label{fig:raw-prediction-error}
    \end{subfigure}
    \begin{subfigure}[t]{0.49\textwidth}
    \centering
    \includegraphics[width=8cm]{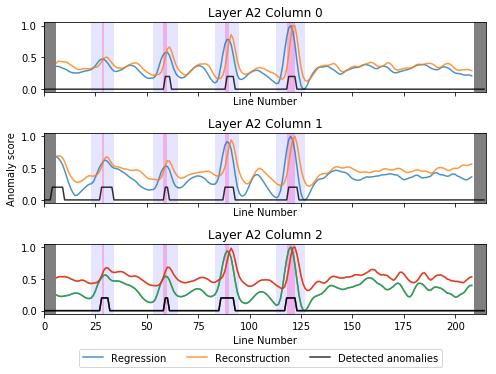}
    \caption{}
    \label{fig:detrended-error}
    \end{subfigure}
\caption{(a) The line-wise reconstruction and regression anomaly scores, averaged on each scan line, and (b) the de-trended and normalized line-wise reconstruction and regression anomaly scores, as well as the detected anomalies. In (a), the raw \ac{MSE} error metric~\eqref{eqn:raw-error} is used to calculate the errors and the anomaly scores; in (b) the spatial scoping error metric~\eqref{eqn:spatial-scoping-loss} is used.  
To give the readers a clearer understanding, we use dark pink shades in (b) to indicate the locations of injected anomalies. A lighter pink color is used to indicate lines adjacent to the injected anomalies if one (or more) window that is used for calculating the anomaly score at this line overlaps with the injected anomalies. In our setting, the affected region has a width of $p+q-1 = 5$ on each side of an anomaly. In the plots, the first (and last) three lines are grayed out to ignore boundary effects. }
\label{fig:error-plots}
\end{figure*}

\begin{figure}[tb]
    \centering
    \includegraphics[width=0.9\linewidth]{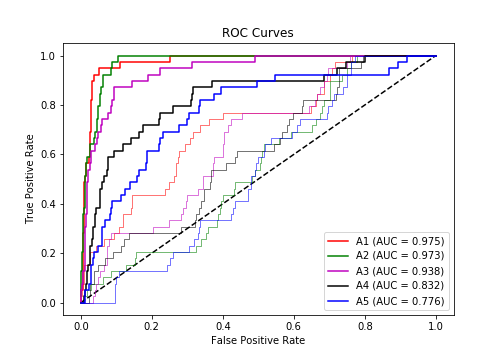}
    \caption{The \ac{ROC} curves of our learning-based model (darker colors) vs. the non-learning model (lighter colors) on the five test layers A1-A5 with injected anomalies.}
    \label{fig:roc}
\end{figure}

\subsection{Data preparation and preprocessing}

\begin{table}[]
\caption{Experiment results on different layers}
\label{tab:results}
\resizebox{\linewidth}{!}{%
\begin{tabular}{ccclc}
\hline
Layers & \begin{tabular}[c]{@{}c@{}}Off-nominal laser power\\ (\% of max value)\end{tabular} & \begin{tabular}[c]{@{}c@{}}Absolute power deviation\\ (\% of max value)\end{tabular} & Precision & Recall \\ \hline
A1 & 58 & 13 & 0.93 & 0.95 \\
A2 & 56 & 11 & 0.90 & 0.99 \\
A3 & 54 & 9 & 0.88 & 0.87 \\
A4 & 50 & 5 & 0.81 & 0.65 \\
A5 & 48 & 3 & 0.75 & 0.61 \\ \hline
\end{tabular}%
}
\end{table}

We choose five nominal layers as our training set and five off-nominal layers with different size of anomalies as our test set; see Table~\ref{tab:results} for details about the off-nominal layers.

In our data preprocessing step, the data was normalized by changing the range (the difference between max and min values) of the data to one. We down-sampled the original $64\times 64$ image data to a resolution of $32\times32$, to reduce the complexity of our network. We chose $p=q=3$ in our experiment for creating the snippets as described in Sec.~\ref{sec:network-design}.

To improve the robustness of our model against small perturbation in input data, we augmented our dataset by adding a small Gaussian noise (with zero mean and a standard deviation of $0.01$\degree C) to the training data. In addition, we know from physics that the thermodynamics of the powder bed is largely governed by the gradient of the temperature distribution; therefore, we can generate additional synthetic data by adding a constant temperature bias $b$ (in our experiment $b\in[-1.8\text{\degree C}, +1.8\text{\degree C}]$) to the original data. This helps regularize our network to better capture the underlying thermodynamics.

\subsection{Network implementation and hyperparameter tuning}
We used Keras~\cite{chollet2015keras} as the framework for implementing the encoder-decoder model. $\lambda$ was set to $1$ so that the reconstruction and regression errors were equally weighted. The model was trained for 500 epochs.

In our unsupervised setting, we did not have labeled anomalous data for using the conventional cross-validation technique to tune the hyper-parameters as in supervised learning. To create a validation set that contains both normal and anomalous data points, we used a simple method to create some synthetic anomalies. $20\%$ of the training data was first picked out as the validation data. We tuned the hyper-parameters (e.g.,~dropout rates and the number of convolution kernels) until the validation loss on normal data had converged.

\subsection{Result analysis and evaluation}

To assess the performance of our model, we tested our model using layers that have different degrees of anomaly. We plotted the distributions of the reconstruction anomaly score in Fig.~\ref{fig:top-down-reconstruction-error} and the regression anomaly score in Fig.~\ref{fig:top-down-regression-error} from a top-down view, with the raw \ac{MSE}~\eqref{eqn:raw-error} used as the error metric. It can be clearly seen that some of the regions with a high anomaly score are located along certain scan lines. To better visualize and quantify the variation of anomaly scores across different scan lines, we display in Fig.~\ref{fig:raw-prediction-error} the average anomaly score along each scan line. From the plot, we can see that scan lines with high anomaly scores appear either as ``sharp peaks'', or as ``big bumps'', which may indicate different causes of anomalous conditions, and should thereby be treated and analyzed separately. 

\vspace{5mm}
\noindent\textbf{Sharp peaks} 
To isolate these sharp peaks in Fig.~\ref{fig:raw-prediction-error}, a detrend technique can be applied to filter out the slowly varying component; here we use a simple detrend technique by subtracting the signal using a window size of $20$. In addition, we also apply the spatial scoping technique~\eqref{eqn:spatial-scoping-loss} as the error metric for calculating the anomaly score. The resulting signal is displayed in Fig.~\ref{fig:detrended-error}, where we can see a clear correlation between the large peaks and the injected anomalies.
A simple thresholding method is used to do anomaly detection. We test our model on the entire training sets and choose the maximum line-wise regression score as the threshold, so no false positives can be detected on all these normal layers. Most injected anomalies can now be correctly detected. We also observe that the first anomaly is difficult to detect. Due to the fact that the deviated laser power merely lasts for 1 line, the temperature there has not yet been significantly changed since the energy accumulated is not sufficient to cause temperature variation in case of limited lasting time and these anomalies may sometimes be buried in the noise. Table~\ref{tab:results} shows the performance of our detection model on the five off-nominal layers. The detection model performs better on layers with higher laser power deviation; the precision and recall rates both exceed $90\%$ in layers A1 and A2. In layers A4 and A5, the precision and recall rates drop significantly, which is due to the reduced temperature disturbance due to smaller laser power deviation. It can also be seen in Fig.~\ref{fig:roc}, the \ac{AUC} rate exceeds 0.97 in layer A1 and layer A2 but drops to only 0.776 in layer A5.

For comparison, we also implemented a simple non-learning method that simply used raw temperature measurement for detecting anomalies. The anomaly score for each image $S_\tau$ is defined as its root-mean-square intensity (temperature) value over the image, i.e.
\begin{align}
    \Score^\text{temp}(\tau) = \Vert{S_\tau}\Vert_F.
\end{align}
As an example, we visualize the distribution of the anomaly scores on Layer A2 from the non-learning method in Fig.~\ref{fig:top-down-temperature}. The \ac{ROC} curves obtained from applying this non-learning method on these anomalous layers are displayed in Fig.~\ref{fig:roc}. It can be seen that our encoder-decoder approach gives a much superior performance to the non-learning method.

\vspace{5mm}
\noindent\textbf{Big bumps}
Having found the cause of sharp peaks, we would like to identify the cause of large bumps in the anomaly scores. By comparing the three top-down views in Fig.~\ref{fig:top-down}, we observe an obvious correlation among these large bump regions. We conjecture that the elevated anomaly scores are due to the high temperature (generally 2\degree C higher than surroundings) and the steep temperature gradient in these parts of the powder bed.

\section{Conclusion}\label{sec:conclusion}
In this paper, we proposed an unsupervised deep learning approach for detecting potential anomalies in an \ac{AM} system. As future work, we plan to apply the proposed technique to other industrial applications. We also plan to conduct a more in-depth theoretical analysis of the proposed technique.


\bibliographystyle{plain}
\bibliography{refs}

\begin{thebibliography}{10}

\bibitem{tofail_review}
Syed A.M.~Tofail, Elias~P. Koumoulos, Amit Bandyopadhyay, Susmita Bose, Lisa
  O'Donoghue, and C.A. Charitidis.
\newblock Additive manufacturing: Scientific and technological challenges,
  market uptake and opportunities.
\newblock {\em Materials Today}, 21, 07 2017.

\bibitem{chandola2009anomaly}
Varun Chandola, Arindam Banerjee, and Vipin Kumar.
\newblock Anomaly detection: A survey.
\newblock {\em ACM computing surveys (CSUR)}, 41(3):15, 2009.

\bibitem{cho2014learning}
Kyunghyun Cho, Bart Van~Merri{\"e}nboer, Caglar Gulcehre, Dzmitry Bahdanau,
  Fethi Bougares, Holger Schwenk, and Yoshua Bengio.
\newblock Learning phrase representations using rnn encoder-decoder for
  statistical machine translation.
\newblock {\em arXiv preprint arXiv:1406.1078}, 2014.

\bibitem{chollet2015keras}
Fran\c{c}ois Chollet et~al.
\newblock Keras.
\newblock \url{https://keras.io}, 2015.

\bibitem{heller2003one}
Katherine Heller, Krysta Svore, Angelos~D Keromytis, and Salvatore Stolfo.
\newblock One class support vector machines for detecting anomalous windows
  registry accesses.
\newblock 2003.

\bibitem{huang_review}
Yong Huang, Ming Leu, Jyoti Mazumder, and M~Donmez.
\newblock Additive manufacturing: Current state, future potential, gaps and
  needs, and recommendations.
\newblock {\em Journal of Manufacturing Science and Engineering}, 137:014001,
  02 2015.

\bibitem{Ioffe2015BatchNA}
Sergey Ioffe and Christian Szegedy.
\newblock Batch normalization: Accelerating deep network training by reducing
  internal covariate shift.
\newblock {\em ArXiv}, abs/1502.03167, 2015.

\bibitem{jin2019one}
Baihong Jin, Yuxin Chen, Dan Li, Kameshwar Poolla, and Alberto
  Sangiovanni-Vincentelli.
\newblock A one-class support vector machine calibration method for time series
  change point detection.
\newblock {\em arXiv preprint arXiv:1902.06361}, 2019.

\bibitem{jin2019detecting}
Baihong Jin, Dan Li, Seshadhri Srinivasan, See-Kiong Ng, Kameshwar Poolla,
  et~al.
\newblock Detecting and diagnosing incipient building faults using uncertainty
  information from deep neural networks.
\newblock {\em arXiv preprint arXiv:1902.06366}, 2019.

\bibitem{krizhevsky2012imagenet}
Alex Krizhevsky, Ilya Sutskever, and Geoffrey~E Hinton.
\newblock {ImageNet} classification with deep convolutional neural networks.
\newblock In {\em Advances in neural information processing systems}, pages
  1097--1105, 2012.

\bibitem{lecun2015deep}
Yann LeCun, Yoshua Bengio, and Geoffrey Hinton.
\newblock Deep learning.
\newblock {\em nature}, 521(7553):436, 2015.

\bibitem{lee2012anomaly}
Yuh-Jye Lee, Yi-Ren Yeh, and Yu-Chiang~Frank Wang.
\newblock Anomaly detection via online oversampling principal component
  analysis.
\newblock {\em IEEE transactions on knowledge and data engineering},
  25(7):1460--1470, 2012.

\bibitem{li2016data}
Dan Li, Guoqiang Hu, and CostasJ Spanos.
\newblock A data-driven strategy for detection and diagnosis of building
  chiller faults using linear discriminant analysis.
\newblock {\em Energy and Buildings}, 128:519--529, 2016.

\bibitem{li2016fault}
Dan Li, Yuxun Zhou, Guoqiang Hu, and Costas~J Spanos.
\newblock Fault detection and diagnosis for building cooling system with a
  tree-structured learning method.
\newblock {\em Energy and Buildings}, 127:540--551, 2016.

\bibitem{malhotra2016lstm}
Pankaj Malhotra, Anusha Ramakrishnan, Gaurangi Anand, Lovekesh Vig, Puneet
  Agarwal, and Gautam Shroff.
\newblock {LSTM}-based encoder-decoder for multi-sensor anomaly detection.
\newblock {\em arXiv preprint arXiv:1607.00148}, 2016.

\bibitem{medel2016anomaly}
Jefferson~Ryan Medel and Andreas Savakis.
\newblock Anomaly detection in video using predictive convolutional long
  short-term memory networks.
\newblock {\em arXiv preprint arXiv:1612.00390}, 2016.

\bibitem{phillips_thesis}
Tim Phillips.
\newblock In-situ laser control method for polymer selective laser sintering
  (sls).
\newblock Master's thesis, University of Texas at Austin, 2016.

\bibitem{phua2010comprehensive}
Clifton Phua, Vincent Lee, Kate Smith, and Ross Gayler.
\newblock A comprehensive survey of data mining-based fraud detection research.
\newblock {\em arXiv preprint arXiv:1009.6119}, 2010.

\bibitem{sakurada2014anomaly}
Mayu Sakurada and Takehisa Yairi.
\newblock Anomaly detection using autoencoders with nonlinear dimensionality
  reduction.
\newblock In {\em Proceedings of the MLSDA 2014 2nd Workshop on Machine
  Learning for Sensory Data Analysis}, page~4. ACM, 2014.

\bibitem{scholkopf2001estimating}
Bernhard Sch{\"o}lkopf, John~C Platt, John Shawe-Taylor, Alex~J Smola, and
  Robert~C Williamson.
\newblock Estimating the support of a high-dimensional distribution.
\newblock {\em Neural computation}, 13(7):1443--1471, 2001.

\bibitem{Simonyan2015VeryDC}
Karen Simonyan and Andrew Zisserman.
\newblock Very deep convolutional networks for large-scale image recognition.
\newblock {\em CoRR}, abs/1409.1556, 2015.

\bibitem{srivastava2015unsupervised}
Nitish Srivastava, Elman Mansimov, and Ruslan Salakhudinov.
\newblock Unsupervised learning of video representations using {LSTM}s.
\newblock In {\em International conference on machine learning}, pages
  843--852, 2015.

\bibitem{vincent2008extracting}
Pascal Vincent, Hugo Larochelle, Yoshua Bengio, and Pierre-Antoine Manzagol.
\newblock Extracting and composing robust features with denoising autoencoders.
\newblock In {\em Proceedings of the 25th international conference on Machine
  learning}, pages 1096--1103. ACM, 2008.

\end{thebibliography}

\end{document}